\documentclass[10pt,letterpaper, conference]{IEEEtran}
\pdfoutput=1
\usepackage{amsmath,graphicx} 
\usepackage{amsfonts, amssymb, amsthm, array}
\usepackage[T1]{fontenc}    
\usepackage{url}            
\usepackage{booktabs}       
\usepackage{nicefrac}       
\usepackage{microtype}      
\usepackage{bbm}
\usepackage{cite}
\newcommand{ \mb}[1]{\mathbf{#1}}

\newcommand{ \mc}[1]{\mathcal{#1}}
\usepackage{epsfig}
\usepackage{multirow}
\newtheorem{theorem}{Theorem}[]

\usepackage{color}
\usepackage{subcaption}
\usepackage[font=small,labelfont=bf]{caption}
\usepackage{algorithmic}
\usepackage{tabularx, booktabs}
\newcolumntype{P}[1]{>{\centering\arraybackslash}p{#1}}
\newcolumntype{Y}{>{\centering\arraybackslash}m{0.9cm}}
\newcolumntype{G}{>{\bfseries\centering\arraybackslash}m{1.8cm}}
\usepackage{etoolbox}{\tiny }
\makeatletter
\@tempswafalse
\@ifpackageloaded{amsthm}{\@tempswatrue}{}
\@ifpackageloaded{ntheorem}{\@tempswatrue}{}
\if@tempswa
  \patchcmd\@thm{\trivlist}{\normalsize\trivlist}{}{}
\else
  \patchcmd\@begintheorem{\trivlist}{\normalsize\trivlist}{}{}
  \patchcmd\@opargbegintheorem{\trivlist}{\normalsize\trivlist}{}{}
\fi
\makeatother

\makeatletter
\g@addto@macro \small {%
 \setlength\abovedisplayskip{5pt plus 2pt minus 2pt}%
 \setlength\belowdisplayskip{5pt plus 2pt minus 2pt}%
}
\makeatother

\makeatletter
\g@addto@macro \footnotesize {%
 \setlength\abovedisplayskip{5pt plus 3pt minus 3pt}%
 \setlength\belowdisplayskip{5pt plus 3pt minus 3pt}%
}
\makeatother

\makeatletter
\g@addto@macro \normalsize {%
 \setlength\abovedisplayskip{5pt plus 3pt minus 3pt}%
 \setlength\belowdisplayskip{5pt plus 3pt minus 3pt}%
}
\makeatother

\newcommand{\edit}[1]{\textcolor{black}{#1}}
\theoremstyle{definition}
\newtheorem{definition}{Definition D.\ignorespaces}[]

\newtheorem{assumption}{Assumption A.\ignorespaces}[]

\newtheorem*{remark*}{Remark}

\newcommand{\startcompact}[1]{\par\vspace{-1em}\begin{#1}%
\allowdisplaybreaks\ignorespaces}

\newcommand{\stopcompact}[1]{\end{#1}\ignorespaces}
\IEEEoverridecommandlockouts

\usepackage{hyperref}
\title{Target--Based Hyperspectral Demixing via Generalized Robust PCA}
\author{\IEEEauthorblockN{Sirisha Rambhatla, Xingguo Li, and Jarvis Haupt}
	\IEEEauthorblockA{Department of Electrical and Computer Engineering,\\ University of Minnesota -- Twin Cities, Minneapolis, MN-55455\\
		{\tt \{rambh002, lixx1661, jdhaupt\}@umn.edu}. }\\
	\vspace*{-0.4in}}

\begin{document}
%
\maketitle
\begin{abstract}
Localizing targets of interest in a given hyperspectral (HS) image has applications ranging from remote sensing to surveillance. This task of target detection leverages the fact that each material/object possesses its own characteristic spectral response, depending upon its composition. As \textit{signatures} of different materials are often correlated, matched filtering based approaches may not be appropriate in this case. In this work, we present a technique to localize targets of interest based on their spectral signatures. We also present the corresponding recovery guarantees, leveraging our recent theoretical results. To this end, we model a HS image as a superposition of a low-rank component and a dictionary sparse component, wherein the dictionary consists of the \textit{a priori} known characteristic spectral responses of the target we wish to localize. Finally, we analyze the performance of the proposed approach via experimental validation on real HS data for a classification task, and compare it with related techniques.
\end{abstract}
\section{Introduction}
\label{sec:intro}
Hyperspectral (HS) imaging is an imaging modality which senses the intensities of the reflected electromagnetic waves corresponding to different wavelengths of the electromagnetic spectra, often invisible to the human eye. As the spectral response associated with an object/material is dependent on its composition, HS imaging lends itself very useful in identifying \edit{target} objects/materials via their characteristic spectra or  \textit{signature} responses, often referred to as "endmembers" in the literature. 
Typical applications of HS imaging range from monitoring agricultural use of land, catchment areas of rivers and water bodies, food processing and surveillance, to detecting various minerals, chemicals, and even presence of life sustaining compounds on distant planets. However, often, these spectral \textit{signatures} are highly correlated, making it difficult to detect region of interest based on these endmembers. Analysis of these characteristic responses to \edit{localize} a \edit{target} material/object serves as the primary motivation of this work. 
\subsection{Model}
\begin{figure}[h]
  \centering
  \begin{tabular}{c}
    \includegraphics[width=0.3\textwidth]{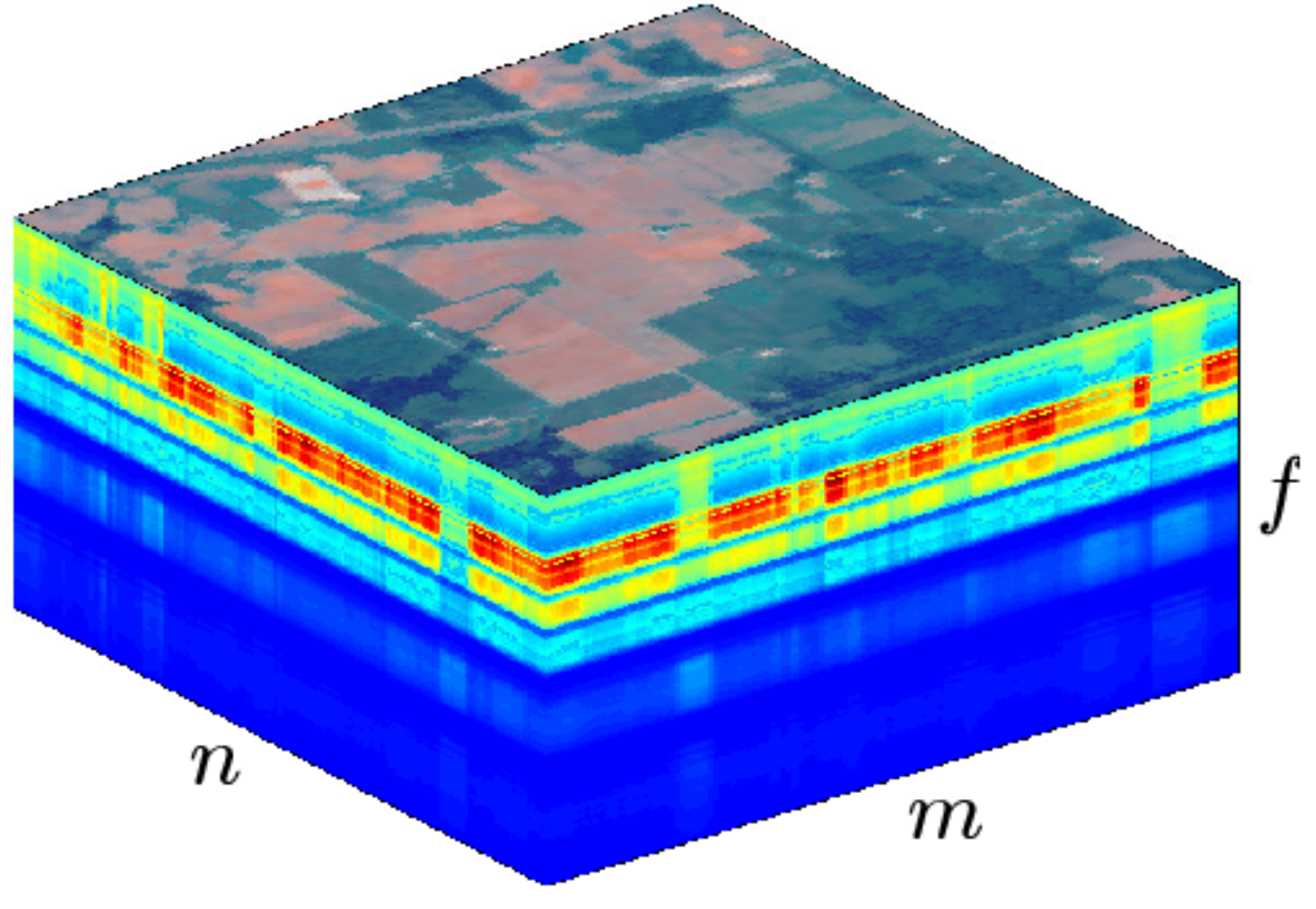}
    \end{tabular}
  \caption{ The Hyperspectral image data-cube corresponding to the Indian Pines dataset. The top layer is depicted in RGB for visualization, however each slice is a grayscale image of reflectances. }
  \label{fig:data} 
\end{figure}

We begin by formalizing the problem. Each HS image {\small $\mb{I} \in \mathbb{R}^{n \times m \times f}$}, is a stack of  {\small$f$} 2-D images each of size {\small$n \times m$}, as shown in Fig.~\ref{fig:data}(a). Here,  {\small$f$} is determined by the number of frequencies or frequency bands across which we measure the reflectances. Overall, each volumetric element or \textit{voxel} , of a HS image is a vector of length {\small $f$}, and represents the response of the material in a pixel of the 2-D grid to {\small $f$} frequencies. 

As a particular scene is composed of only a small number of objects/materials, the corresponding characteristic responses of each voxel are low-rank \cite{Keshava2002}. For example, while imaging an agricultural area, we would expect to record responses from materials like biomass, farm vehicles, roads, houses and water bodies and so on. Further, the spectra of complex materials can be assumed to be a linear mixture of the constituent materials \cite{Keshava2002, Greer2012}, i.e. HS image voxels can be viewed as being generated by a linear mixture model \cite{Xing2012}. Since, a particular scene does not contain a large number of distinct materials, we can decompose the responses into a low-rank part and a component which is sparse in a known dictionary -- a \textit{dictionary sparse} part. 

Formally, let {\small$\mb{Y}\in\mathbb{R}^{ f \times nm}$} be formed by \textit{unfolding} the HS image {\small$\mb{I}$}, such that, each column of {\small$\mb{Y}$} corresponds to a voxel. Now, let {\small$\mb{Y}$} arise as a result of a superposition of a low-rank component {\small$\mb{X} \in \mathbb{R}^{f \times nm}$} with rank {\small $r$}, and a dictionary sparse component expressed here as {\small$\mb{RA}$}, i.e.,
	\startcompact{small}
	\begin{align}
	\label{Prob}
	\mb{Y} = \mb{X} + \mb{RA}.
	\end{align}
	\stopcompact{small}%
Here, {\small$\mb{R} \in \mathbb{R}^{f \times d}$} is an \textit{a priori} known dictionary composed of normalized characteristic responses of the material/object, \edit{or of the constituents of the material} we wish to \edit{localize}, and {\small$\mb{A} \in \mathbb{R}^{d \times nm}$} is the \textit{sparse} coefficient matrix (also referred to as "abundances" in the literature) with {\small $s$} non-zero elements globally, i.e. {\small$\|\mb{A}\|_0 =s$}. Note that, {\small$\mb{R}$} can also be constructed by learning a dictionary based on the known spectral signatures of a target\cite{Mairal10,Olshausen97,Aharon05}. 

\subsection{Prior Art}
\label{priorart}
The model shown in \eqref{Prob} is a close counterpart of a number of well-known problems. To start, in the absence of the dictionary sparse part, \eqref{Prob} reduces to the popular problem of principal component analysis (PCA) \cite{Jolliffe02}. This problem also shares its structure with variants of PCA such as robust-PCA\cite{Candes11, Chandrasekaran11} where  {\small$\mb{R}$} is absent, outlier pursuit \cite{Xu2010} where {\small$\mb{R} = \mb{I}$} and {\small$\mb{A}$} is column sparse, and others  \cite{Zhou2010, Ding11, Wright13, Chen13, Li15, Li15c, Li15b, Li16_refined, Li2016efficient}. On the other hand, the problem can be identified as that of sparse recovery \cite{Natarajan95, Donoho01, Candes05}, in the absence of the low-rank part {\small$\mb{X}$}. Following which, sparse recovery methods for analysis of HS images have been explored in \cite{Moudden2009, Bobin2009, Kawakami2011, Charles2011}. In addition, in a recent work \cite{Giampouras2016}, the authors further impose a low-rank constraint on the coefficient matrix {\small$\mb{A}$}. Further, applications of compressive sampling have been explored in \cite{Golbabaee2010}, while \cite{Xing2012} analyzes the case where HS images are noisy and incomplete.

The model described in \eqref{Prob} was introduced in \cite{Mardani2012} as a means to detect traffic anomalies in a network, wherein, the authors focus on a case where the dictionary {\small$\mb{R}$} is \textit{overcomplete}, i.e., \textit{fat}, and the rows of {\small$\mb{R}$} are orthogonal, e.g., {\small$\mb{RR^\top} = \mb{I}$}. Further, the coefficient matrix {\small$\mb{A}$} is assumed to possess at most {\small $k$} nonzero elements per row and column and {\small $s$} nonzero elements overall. 

In a recent work \cite{Rambhatla2016}, we analyze the extension of \cite{Mardani2012} to a case where the dictionary has more rows than columns, i.e., is \textit{thin}, while removing the orthogonality constraint for both the \textit{thin} and the \textit{fat} dictionary cases, when {\small$s$} is small. The \textit{thin} dictionary case is amenable to a number of applications, as often, we are looking for only a small number of \textit{signatures}. For example, in our motivating application of HS imaging, we are interested in detecting the presence of a small number of spectral \textit{signatures} corresponding to the target of interest. To this end, we focus our attention on the \textit{thin} case, although a similar analysis applies for the \textit{fat} case \cite{Rambhatla2016}. 

Note that, when the known dictionary {\small $\mb{R}$} is thin, we can alternatively multiply \eqref{Prob} on the left by the pseudo-inverse of {\small $\mb{R}$}, say {\small $\mb{R}^\dagger$}, in which case, the model shown in \eqref{Prob} reduces to that of robust PCA, i.e.,
	\startcompact{small}
	\begin{align}
	\label{Prob_rpca}
	\mb{\tilde{Y}} = \mb{\tilde{X}} + \mb{A},
	\end{align}
	\stopcompact{small}%
where {\small $	\mb{\tilde{Y}} = \mb{R}^\dagger\mb{Y}$} and {\small $	\mb{\tilde{X}} = \mb{R}^\dagger\mb{X}$}. Therefore, in this case, we can adopt a two-step procedure: (1) recover the sparse matrix {\small $\mb{A}$} by robust PCA and, (2) estimate the low-rank part using the estimate of {\small $\mb{RA}$}. \footnote{We denote this method of estimating {\footnotesize $\mb{A}$} as \texttt{RPCA$^\dagger$} in our experiments.} Models considered in \eqref{Prob} and \eqref{Prob_rpca} are two alternate methods of recovering the components for thin dictionaries. These two methods inherently solve different optimization problems. The proposed method is a one-shot procedure with guarantees to solve the problem, while the conditions under which the pseudo-inverse based procedure succeeds will depend on the interaction between {\small $\mb{R}^\dagger$} and the low-rank part. Further, our experiments indicate that solving for model in \eqref{Prob} is more \textit{robust} as compared to \texttt{RPCA$^\dagger$} for the classification problem at hand, across different choice of dictionaries. 
\subsection{Our Contribution}
In this work, we present a one-step method for target detection in a hyperspectral image with the underlying model of \eqref{Prob}. This is done by building upon the results of \cite{Rambhatla2016} for the underlying application. In particular, we show the conditions under which target detection is possible with thin dictionaries. In general, the method is also applicable to fat dictionaries and for other applications, which follow the underlying model. The rest of the paper is organized as follows. We formulate the problem, introduce the notation and present the theoretical result for the case at hand, i.e., the \textit{thin} case, in section~\ref{sec:Formulation}. Next, we present the description of the dataset along with the experimental evaluations in section~\ref{sec:exp}. Finally, we conclude this discussion in section~\ref{sec:conclusion}.
	\section{Problem Formulation}
	\label{sec:Formulation}	
 	Our aim is to recover the low-rank component {\small$\mb{X}$}, and the sparse coefficient matrix {\small$\mb{A}$}, given the dictionary {\small$\mb{R}$}, and samples {\small$\mb{Y}$} generated according to the model described in \eqref{Prob}. We utilize the structure of the components, {\small$\mb{X}$} and {\small$\mb{A}$}, to arrive at the following convex problem for {\small$\lambda \geq 0 $},
	\startcompact{small}
	\begin{align}
	\label{optProb}
	\underset{\mb{X}, \mb{A}}{\text{minimize~}} \|\mb{X}\|_* + \lambda \|\mb{A}\|_1 ~~\text{s.t.}~~ \mb{Y} = \mb{X} + \mb{RA}.
	\end{align}
	\stopcompact{small}%
	where, {\small$\|.\|_* $} denotes the nuclear norm, and {\small$\|.\|_1$} refers to the $l_1$- norm, which serve as convex relaxations of rank and sparsity (i.e. $l_0$-norm), respectively. For the application at hand, we present the theoretical result for the case when the number of dictionary elements {\small $d \leq f$}. Analysis of a more general case can be found in \cite{Rambhatla2016}. For the this case, we assume the dictionary {\small$\mb{R}$} to be a \textit{frame}, i.e. for any vector $\mb{v}\in \mathbb{R}^{d}$, we have
	\startcompact{small}
	\vspace{-5pt}
	\begin{align}
	\label{frame}
	\mb{F}_{L}\|\mb{v}\|^2_2 \leq \|\mb{Rv}\|^2 \leq \mb{F}_U\|\mb{v}\|^2_2, 
	\end{align}
	\stopcompact{small}%
	where {\small$\mb{F}_{L}$} and {\small$\mb{F}_U$} are the lower and upper \textit{frame bounds}, respectively, with {\small$0<\mb{F}_{L}\leq\mb{F}_U$}. 

	We begin by defining a few relevant subspaces, similar to those used in \cite{Mardani2012}.

	\begin{table*}[th]
	\caption{Simulation results  for the proposed approach ({\footnotesize $\mb{XpRA}$}), robust-PCA based approach (\texttt{RPCA$^\dagger$}), matched filtering (\texttt{MF}) on original data {\footnotesize $\mb{Y}$}, and matched filtering on transformed data {\footnotesize $\mb{R^\dagger Y}$} (\texttt{MF$^\dagger$}), across dictionary elements $d$, and the regularization parameter for initial dictionary learning step $\rho$. Threshold selects columns with column-norm greater than threshold such that AUC is maximized. Further, $`` * "$ denotes the case where ROC curve was ``flipped'' (i.e. classifier output was inverted to improve performance). }

	\label{res_tab}
\captionsetup{justification=centering}
	\begin{subtable}{.35\linewidth}
	\captionsetup{font=footnotesize}
	 \centering
	 \caption{Learned dictionary, $d=4$}
   	\scalebox{0.72}{
   	\begin{tabular}{|P{1cm}|c|c|c|c|c|c|}
	\hline
	\multirow{2}{*}{\textbf{$d$}} & \multirow{2}{*}{$\rho$} & \multirow{2}{*}{\textbf{Method}}& \multirow{2}{*}{\textbf{Threshold }}&\multicolumn{2}{G|}{\textbf{Performance at best operating point} }& \multirow{2}{*}{\textbf{AUC}}\\ \cline{5-6}
	&&&&\textbf{TPR}&\textbf{FPR}&\\\hline
	\multirow{12}{*}{4} &	\multirow{4}{*}{0.01}
	  &\textbf{XpRA}					&0.30&\textbf{0.979}&\textbf{0.023}&\textbf{0.989}\\\cline{3-7}
	&&\textbf{RPCA$^\dagger$} &0.65&0.957&0.049&0.974\\\cline{3-7}
	&&\textbf{MF$_*$}						&N/A&0.957&0.036&0.994\\\cline{3-7}
	&&\textbf{MF$_*^\dagger$}	   &N/A&0.914&0.104&0.946\\\cline{2-7}
	&	\multirow{4}{*}{0.1}
  &\textbf{XpRA}				    	&0.8&\textbf{0.989}&0.017&0.997\\\cline{3-7}
	&&\textbf{RPCA$^\dagger$} &0.8&\textbf{0.989}&0.014&0.997\\\cline{3-7}
	&&\textbf{MF}						&N/A&\textbf{0.989}&0.016&\textbf{0.998}\\\cline{3-7}
	&&\textbf{MF$^\dagger$}	   &N/A&\textbf{0.989}&\textbf{0.010}&\textbf{0.998}\\\cline{2-7}
	&	\multirow{4}{*}{0.5}
  &\textbf{XpRA}					    &0.6&\textbf{0.968}&\textbf{0.031}&\textbf{0.991}\\\cline{3-7}
	&&\textbf{RPCA$^\dagger$} &0.6&0.935&0.067&0.988\\\cline{3-7}
	&&\textbf{MF}						&N/A&0.548&0.474&0.555\\\cline{3-7}
	&&\textbf{MF$_*^\dagger$}	   &N/A&0.849&0.119&0.939\\\hline
	\end{tabular}
	\label{dl_4}
	}
	\end{subtable} 
	\begin{subtable}{.35\linewidth}
	\captionsetup{font=footnotesize}
	 \centering
	 \caption{Learned dictionary, $d=10$}
	   	\scalebox{0.72}{
	   	\begin{tabular}{|P{1cm}|c|c|c|c|c|c|}
		\hline
		\multirow{2}{*}{\textbf{$d$}} & \multirow{2}{*}{$\rho$} & \multirow{2}{*}{\textbf{Method}}& \multirow{2}{*}{\textbf{Threshold}}&\multicolumn{2}{G|}{\textbf{Performance at best operating point} }& \multirow{2}{*}{\textbf{AUC}}\\ \cline{5-6}
			&&&&\textbf{TPR}&\textbf{FPR}&\\\hline
		\multirow{12}{*}{10} &	\multirow{4}{*}{0.01}
		  &\textbf{XpRA}					&0.6&0.935&0.060&0.972\\\cline{3-7}
		&&\textbf{RPCA$^\dagger$} &0.7&\textbf{0.978}&\textbf{0.023}&\textbf{0.990}\\\cline{3-7}
		&&\textbf{MF$_*$}						&N/A&0.624&0.415&0.681\\\cline{3-7}
		&&\textbf{MF$^\dagger_*$}	   &N/A&0.569&0.421&0.619\\\cline{2-7}
		&	\multirow{4}{*}{0.1}
	  &\textbf{XpRA}					    &0.5&\textbf{0.968}&\textbf{0.029}&\textbf{0.993}\\\cline{3-7}
		&&\textbf{RPCA$^\dagger$} &0.5&0.871&0.144&0.961\\\cline{3-7}
		&&\textbf{MF$_*$}						&N/A&0.688&0.302&0.713\\\cline{3-7}
		&&\textbf{MF$^\dagger$}	   &N/A&0.527&0.469&0.523\\\cline{2-7}
		&	\multirow{4}{*}{0.5}
	  &\textbf{XpRA}					    &1   &\textbf{0.978}&\textbf{0.031}&\textbf{0.996}\\\cline{3-7}
		&&\textbf{RPCA$^\dagger$} &2.2&0.849&0.113&0.908\\\cline{3-7}
		&&\textbf{MF}						&N/A&0.807&0.309&0.781\\\cline{3-7}
		&&\textbf{MF$^\dagger_*$}	   &N/A&0.527&0.465&0.539\\\hline
		\end{tabular}
		\label{dl_10}
		}
		\end{subtable}
		\begin{subtable}{.27\linewidth}
		\captionsetup{font=footnotesize}
		 \centering
		 \caption{Dictionary by sampling voxels, $d=15$}
			   	\scalebox{0.66}{
			   	\begin{tabular}{|P{1cm}|c|c|c|c|c|}
				\hline 
				\multirow{2}{*}{\textbf{$d$}}  & \multirow{2}{*}{\textbf{Method}}& \multirow{2}{*}{\textbf{Threshold}}&\multicolumn{2}{G|}{\textbf{Performance at best operating point} }& \multirow{2}{*}{\textbf{AUC}}\\ \cline{4-5}
				&&&TPR&FPR&\\\hline
				\multirow{4}{*}{15}
				  &\textbf{XpRA}					&0.3&\textbf{0.989}&\textbf{0.021}&\textbf{0.998}\\\cline{2-6}
				&\textbf{RPCA$^\dagger$} &3&0.849&0.146&0.900\\\cline{2-6}
				&\textbf{MF}						&N/A&0.957&0.085&0.978\\\cline{2-6}
				&\textbf{MF$^\dagger$}	   &N/A&0.796&0.217&0.857\\\hline
				\end{tabular}
				\label{orig_dict}
				}
					 \vspace{0.17cm}
				\caption{Average performance}
					 \vspace{-3px}
				\scalebox{0.61}{	
				\begin{tabular}{|P{1cm}|c|c|c|c|c|c|}		
				\hline
				\multirow{2}{*}{\textbf{Method}}& \multicolumn{2}{G|}{\textbf{TPR}}& \multicolumn{2}{G|}{\textbf{FPR}}& \multicolumn{2}{G|}{\textbf{AUC}}\\ \cline{2-7}
				&Mean&St.Dev.&Mean&St.Dev.&Mean&St.Dev.\\ \hline
				\textbf{XpRA}&	\textbf{0.972}&	\textbf{0.019}&	\textbf{0.030}&	\textbf{0.014}&	\textbf{0.991}&	\textbf{0.009}\\ \hline
				\textbf{RPCA$^\dagger$}&0.919&0.061&0.079&0.055&0.959&0.040\\ \hline
				\textbf{MF}&0.796&0.179&0.234&0.187&0.814&0.178\\ \hline
				\textbf{MF$^\dagger$}	&0.739&0.195&0.258&0.192&0.775&0.207\\ \hline
				\end{tabular}
				\label{overall_perf}
				}
				\end{subtable}
	\captionsetup{font=small}
	\vspace{-12pt}
	\end{table*}

	\captionsetup{justification=justified}
\setlength\extrarowheight{5pt}
	Let the pair {\small$\{\mb{X_0, A_0} \}$} be the solution to the problem shown in \eqref{optProb}. We define {\small$\mb{\Phi}$} as the space of matrices spanning either the row or the column space of the low-rank component {\small$\mb{X}_0$}. Specifically, let {\small$\mb{U \Sigma V^\top}$} denote the singular value decomposition of {\small$\mb{X}_0$}, then  the space  {\small$\mb{\Phi}$} is defined as
	\startcompact{small}
	\begin{align*}
	\mb{\Phi} := \{ \mb{UW^\top_1} + \mb{W_2}\mb{V^\top}, \mb{W_1}\in \mathbb{R}^{ nm\times r}, \mb{W_2} \in \mathbb{R}^{f \times r}\}.
	\end{align*}
	\stopcompact{small}%
	Next, let {\small$\mb{\Omega}$} be the space spanned by the matrices which have the same support (locations of non-zero elements) as {\small$\mb{A}_0$}, and let {\small$\mb{\Omega_R}$} be defined as
 	\startcompact{small}
 	\begin{align*}
 	\mb{\Omega_R} := \{ \mb{Z} = \mb{RH}, \mb{H} \in \mb{\Omega} \}.
 	\end{align*}
 	\stopcompact{small}%
 	 Next, let {\small$\mc{P}_\Phi(.)$}, {\small$\mc{P}_\Omega(.)$} and {\small$\mc{P}_{\Omega_R}(.)$} be the orthogonal projection operator(s) onto the space of matrices defined above. In addition, we will use {\small$\mb{P}_U$} and {\small$\mb{P}_V$} to denote the \edit{projection matrices for} the column and row spaces of {\small$\mb{X}_0$}, respectively, i.e., implying the following for any matrix {\small$\mb{X}$},%
    \startcompact{small}
	\begin{align*}
	\mc{P}_\Phi (\mb{X}) &= \mb{P}_U\mb{X} + \mb{X} \mb{P}_V -   \mb{P}_U\mb{X}\mb{P}_V
	\end{align*}
	\stopcompact{small}%
	Indeed, there are situations under which we cannot hope to recover the matrices {\small$\mb{X}$} and {\small$\mb{A}$}. To characterize these scenarios, we employ various notions of incoherence, out of which, the first is the incoherence between the low-rank part, {\small$\mb{X}_0$}, and the dictionary sparse part, {\small$\mb{RA_0}$}, 
	\startcompact{small}
	\begin{align*}
	\mu := \underset{\mb{Z} \in \mb{\Omega}_R \backslash \{\mb{0}_{d \times nm}\}}{\text{max}} \tfrac{\|\mathcal{P}_{\Phi}(\mb{Z})\|_F}{\|\mb{Z}\|_F},
	\end{align*}
	\stopcompact{small}%
	where $\mu \in [0, 1]$ is small when these components are incoherent (which is good for recovery).  The next two measures of incoherence can be interpreted as a way to avoid the cases where for {\small$\mb{X}_0 = \mb{U \Sigma V^\top}$}, (a) {\small$\mb{U}$} resembles the dictionary  {\small$\mb{R}$}, and (b) {\small$\mb{V}$} as the sparse coefficient matrix {\small$\mb{A}_0$}. To this end, we define respectively, the following to measure these properties,
	\startcompact{small}
    \begin{align*}
	\mb{\gamma}_{U\!R}  :=  \underset{i}{\text{max}} \tfrac{\|\mb{P}_U \mb{R}\mb{e}_{i}\|^2}{\|\mb{Re}_{i}\|^2} ~{\normalsize\text{and}}~
	\mb{\gamma}_V := \underset{i}{\text{max}} \|\mb{P}_V\mb{e}_{i}\|^2,
	\end{align*}
	\stopcompact{small}%
	 where {\small$\mb{\gamma}_V  {\small\in [r/nm, 1]}$}. Also, we define {\small$\xi := \|\mb{R^\top}\mb{UV^\top}\|_\infty$}.
%
	\edit{Finally, for a matrix {\small $\mb{M}$}, we} use {\small$\|\mb{M}\|:= 
	\sigma_{\text{max}}(\mb{M})$} for the spectral norm, where {\small$\sigma_{\text{max}}(\mb{M})$} denotes the maximum singular value of the matrix {\small$\mb{M}$}, and {\small$\|\mb{M}\|_\infty := \underset{\{i,~j\}}{\text{max}} |M_{ij}|$}. 
\subsection{Theoretical Result}
\label{sec:Theory}
We now present our main theoretical result for the \textit{thin} case, i.e., when {\small$ d \leq f$}, which was developed in \cite{Rambhatla2016}. We begin by introducing the following definitions and assumptions. 
\begin{definition}
\label{lamMin}
{\small$ \lambda_{\text{min}} = \tfrac{1 + C}{1-C} ~\xi$}, where $C$ is defined as 
\startcompact{small}
\begin{align}
 {C} := \tfrac{c}{\mb{F}_L(1 - \mu)^2 - c}, \notag
 \end{align}
\stopcompact{small}%
{\footnotesize $c := \tfrac{\mb{F}_U}{2} [(1 + 2\gamma_{U\!R} )(\text{min}(s, d)  + s\gamma_V ) +2s\gamma_V ] - \tfrac{\mb{F}_L}{2}[\text{min}(s, d)  + s\gamma_V ]$}.
\end{definition}
\vspace{0.1em}
\begin{definition}%
\label{lamMax}%
\startcompact{footnotesize}%
$\lambda_{\text{max}} := \tfrac{1}{\sqrt{s}} \big(\sqrt{\mb{F}_{L}} {(}~ 1- \mu ~{)} -\sqrt{r \mb{F}_U} \mu\big{)}$.
\stopcompact{footnotesize}%
\end{definition}
\vspace{0.1em}
\begin{assumption}
\label{A1}
\startcompact{small}%
$\lambda_{\text{max}} \geq \lambda_{\text{min}}$
\stopcompact{small}%
\end{assumption}
\vspace{0.1em}
\begin{assumption}
\label{A2}
\startcompact{small}%
Let $s_{\text{max}} := \tfrac{(1 - \mu)^2}{2}\tfrac{nm}{r} $, then
\begin{align*}%
\gamma_{U\!R} ~~\leq ~~
	\begin{cases}
     \frac{(1 - \mu)^2 - 2s\gamma_V}{2s( 1 + \gamma_V)}, \text{ for }    s \leq \text{\small min} ~(d, s_{\text{max}})\\
     \frac{(1 - \mu)^2 - 2s\gamma_V}{2(d + s\gamma_{V})}, \text{ for }   d<s \leq s_{\text{max}} 
     \end{cases}. \notag 
\end{align*}
\stopcompact{small}%
\end{assumption}
\noindent The following theorem establishes the conditions under which we can recover the pair {\small$\{\mb{X_0, A_0} \}$} successfully.
\begin{theorem}
\label{theorem}
Consider a superposition of a low-rank matrix {\small $\mb{X}_0 \in \mathbb{R}^{f \times nm}$} of rank {\small $r$}, and a dictionary sparse component {\small $\mb{RA_0}$}, wherein the dictionary {\small $\mb{R} \in \mathbb{R}^{f \times d}$}, with {\small$d\leq f$}, obeys the frame condition with frame bounds {\small $[\mb{F}_{L}, \mb{F}_U ]$}, the sparse coefficient matrix {\small $\mb{A}_0$} has at most {\small $s$} non-zeros, i.e., {\small $\|\mb{A}_0\|_0 = s$}, where {\small $s \leq s_{\text{max}}$}, and {\small $\mb{Y} = \mb{X}_0 + \mb{RA_0}$}, with parameters {\small$\gamma_{U\!R}$},  {\small $\xi$}, {\small$\mb{\gamma}_V \in [r/nm, 1]$} and  {\small$\mu \in [0, 1]$}. Then, solving the formulation shown in \eqref{optProb} will recover matrices {\small $\mb{X}_0$} and {\small $\mb{A}_0$} if assumptions A.\ref{A1} and A.\ref{A2} hold for any {\small $\lambda \in [\lambda_{\text{min}}, \lambda_{\text{max}} ]$}, as defined in D.\ref{lamMin} and  D.\ref{lamMax}, respectively.
\end{theorem}
\noindent\edit{In other words, Theorem~\ref{theorem} establishes the sufficient conditions for the existence of {\small $\lambda$}s to guarantee recovery of  {\small$\{\mb{X_0, A_0} \}$}. We see that these conditions are closely related to the various incoherence measures {\small$\gamma_{U\!R}$}, {\small$\mb{\gamma}_V$} and {\small$\mu$}, between {\small$ \mb{X}$}, {\small$ \mb{R}$} and {\small$ \mb{A}$} . Note that, we have here an upper-bound on the global sparsity, i.e., $s \leq s_{\text{max}}$, however, our simulation results in \cite{Rambhatla2016} on synthetic data reveal that the recovery is indeed possible even for higher {\small$s$}; See \cite{Rambhatla2016} for the details of the analysis. To this end, we propose a randomized analysis of the problem in \eqref{optProb} to improve these results as a future work.} 
\section{Experimental Evaluation}
\label{sec:exp}
In this section, we evaluate the performance of the proposed technique on real HS data. We begin by describing the dataset used for the simulations, following which we describe the experimental set-up and conclude by presenting the results.
\subsection{Data}
The Airborne Visible/Infrared Imaging Spectrometer (AVIRIS) \cite{AVIRIS} sensor is one of the popular choices for collecting HS images for various remote sensing applications. In this present exposition, we consider the ``Indian Pines'' dataset \cite{HSdat}, which was collected over the Indian Pines test site in North-western Indiana in the June of 1992. This dataset consists of spectral reflectances across $224$ bands in wavelength of ranges $400-2500$ nm from a scene which is composed mostly of agricultural land, along with two major dual lane highways, a rail line and some built structures. This dataset is further processed by removing the bands corresponding to those of water absorption, and the resulting HS data-cube with dimensions $\{145 \times 145 \times 200\}$ is as visualized in Fig.~\ref{fig:data}. This modified dataset is available as ``corrected Indian Pines'' dataset \cite{HSdat}, with the ground-truth containing $16$ classes; Henceforth, referred to as the ``Dataset". 

We will analyze the performance of the proposed technique for the identification of the stone-steel towers (class $16$ in the Dataset), shown as ground-truth in Fig.~\ref{figure:res_our}(b), which constitutes about $93$ voxels in the Dataset.  We consider two cases for choosing the known dictionary {\small$\mb{R}$} -- (1) when a (thin) dictionary is learned\cite{Mairal10,Olshausen97,Aharon05} based on the voxels (class-$16$), and (2) when the dictionary is formed by randomly sampling voxels (from class-$16$). This is to emulate the two ways in which we can arrive at the characteristics of a target -- (1) the case where the \textit{exact signatures} are not available, and/or there is noise, and (2) where we have access to the exact signatures we are looking for. Next, we form the data matrix  {\small$\mb{Y} \in \mathbb{R}^{f \times nm}$} by stacking each voxel of the image side-by-side, which results in a $\{200 \times 145^2\}$ data matrix {\small$\mb{Y}$}. \\
%
%
\\
\noindent\textbf{Normalization:} For normalization, we divide the data matrix {\small$\mb{Y}$} and the dictionary {\small$\mb{R}$} by {\small$\|\mb{Y}\|_{\infty}$} to preserve the inter-voxel scaling. Next, we normalize the columns of {\small$\mb{R}$}, such that they are unit-norm. 
In practice, we can store the un-normalized dictionary  {\small$\mb{R}$}, consisting of actual \textit{signatures} of the target material, and can normalize it, as described above, after the HS image has been acquired.
\subsection{Experimental Setup}
We evaluate and compare the performance of the proposed method (\texttt{XpRA}) with \texttt{RPCA$^\dagger$} using transformed data {\small $\mb{\tilde{Y} = R^\dagger Y}$} (as described in section~\ref{priorart}), matched filtering (MF) based approach on data {\small $\mb{Y}$} ( denoted by \texttt{MF}), and a MF based approach on {\small $\mb{\tilde{Y}}$}, denoted by \texttt{MF$^\dagger$}, via receiver \edit{operating} characteristic (ROC), shown in Table~\ref{res_tab}(a)-(d). 

For the proposed technique, we employ the accelerated proximal gradient (APG) algorithm presented in \cite{Mardani2012} to solve the optimization problem shown in \eqref{optProb}. Similarly, for \texttt{RPCA$^\dagger$} we employ the APG with transformed data matrix {\small $\mb{\tilde{Y}}$}, while setting {\small $\mb{R = I}$}; See \eqref{Prob_rpca}. With reference to selection of tuning parameters for the APG solver for {\texttt{XpRA} (\texttt{RPCA$^\dagger$}, respectively), we choose $v = 0.95$, {\small$\nu =\|\mb{Y}\|$} ({\small$\nu =\|\mb{\tilde{Y}}\|$}), {\small$\bar{\nu} = 10^{-4}$}, and scan through $100$ $\lambda$s in the range {\small$\lambda \in (0, {|\mb{R^\top Y}\|_{\infty}}/{\|\mb{Y}\|} ]$} {\small($\lambda \in (0, {\|\mb{\tilde{Y}}{\|_{\infty}}/{\|\mb{\tilde{Y}}\|} }]$)}, to generate the ROCs. The output of \texttt{XpRA} and  \texttt{RPCA$^\dagger$} is a data-cube corresponding to the low-rank part and the dictionary sparse part (which contains the region of interest).  We threshold the resulting estimates of the sparse part {\small $\mb{A}$}, based on their column norms. We choose the threshold such that the area under the curve (AUC) is maximized for both cases (\texttt{XpRA} and \texttt{RPCA$^\dagger$}). 

In case of \texttt{MF}, we form the inner-product of the \edit{column-normalized} data matrix {\small $\mb{Y}$}, say {\small $\mb{Y}_n$},  with the dictionary {\small $\mb{R}$}, i.e., {\small $\mb{R^\top Y}_n$}, and select the maximum absolute inner-product per column. For \texttt{MF$^\dagger$}, matched filtering corresponds to finding maximum absolute entry for each column of the column-normalized {\small $\mb{\tilde{Y}}$}. Next, in both cases we scan through $1000$ threshold values between $(0, \edit{1}]$ to generate the ROC results shown in Table~\ref{res_tab}. 

Table~\ref{res_tab}(a)-(b) show the ROC characteristics for the aforementioned methods for a dictionary learned using all voxels of class-$16$. We evaluate the performance of the aforementioned techniques for two dictionary sizes, $d=4$ and $d=10$, for three values of the regularization parameter $\rho$, used for the initial dictionary learning step, i.e., $\rho = 0.01,~0.1$ and $0.5$. (The parameter $\rho$ controls the sparsity during the initial dictionary learning step.) 

Table~\ref{res_tab}(c) shows the ROC results for the case where we randomly select $15$ voxel from class-$16$ to form the known dictionary. Further, we show the overall performance in terms of the mean and standard deviation of ROC metrics (TPR, FPR and AUC), for each algorithm in Table~\ref{res_tab}(d), i.e., across different dictionaries considered in Table~\ref{res_tab}(a)-(c). 
\subsection{Analysis}
As described above, Table~\ref{res_tab} shows the ROCs for the classification performance of the proposed method as compared to \texttt{RPCA$^\dagger$}, \texttt{MF} and \texttt{MF$^\dagger$}. We note that when we use {\small $\mb{R}$} which has been learned based on samples of class-$16$, matched filtering based approaches under perform (with an exception for $d=4$ and $\rho=0.1$). \texttt{RPCA$^\dagger$} performs well for $d=10$ and $\rho=0.01$. For all other choices the proposed method performs the best. Even for cases in which it is not the best, it still is comparable to the best performer.  Following this, we conclude that the performance of \texttt{XpRA} is most reliable across different the choices of the dictionaries. This is also evident by Table~\ref{res_tab}(d) which shows the mean and and standard deviation of ROC metrics (TPR, FPR and AUC). We observe that the mean and standard deviation metrics for the proposed method are superior to other methods across the different dictionary choices. 
\begin{figure}[bth]
\centering
\begin{tabular}{cP{0.01cm}cc}
{\footnotesize \textbf{Data}}&&{\footnotesize$\mb{X}$ }& {\footnotesize $\mb{RA}$} \vspace{-2pt}\\
  \epsfig{file=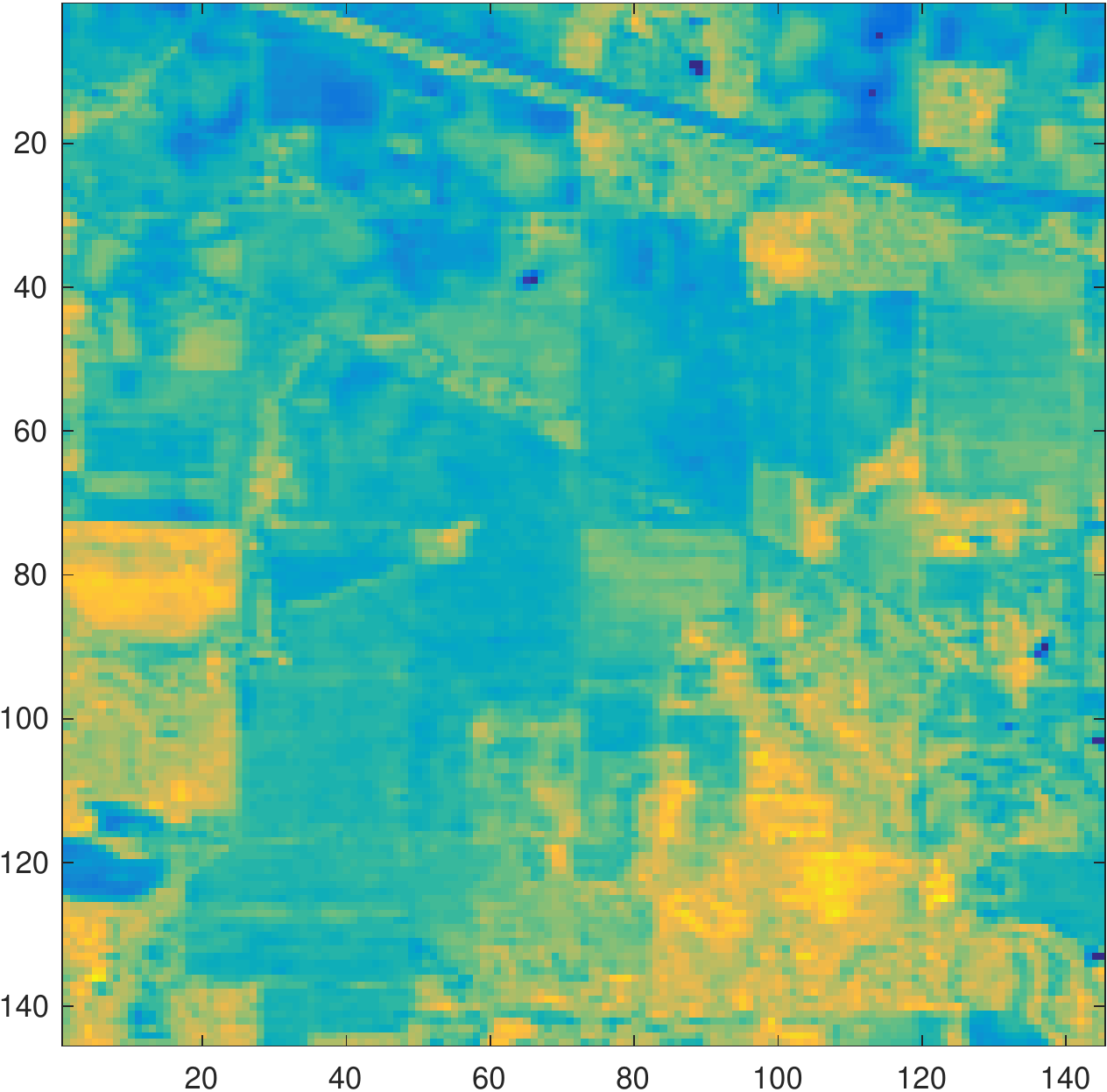,width=0.27\linewidth,clip=}&
   {\small\rotatebox{90}{ ~~~~~~~~~{Best $\lambda$}}} & 
   \epsfig{file=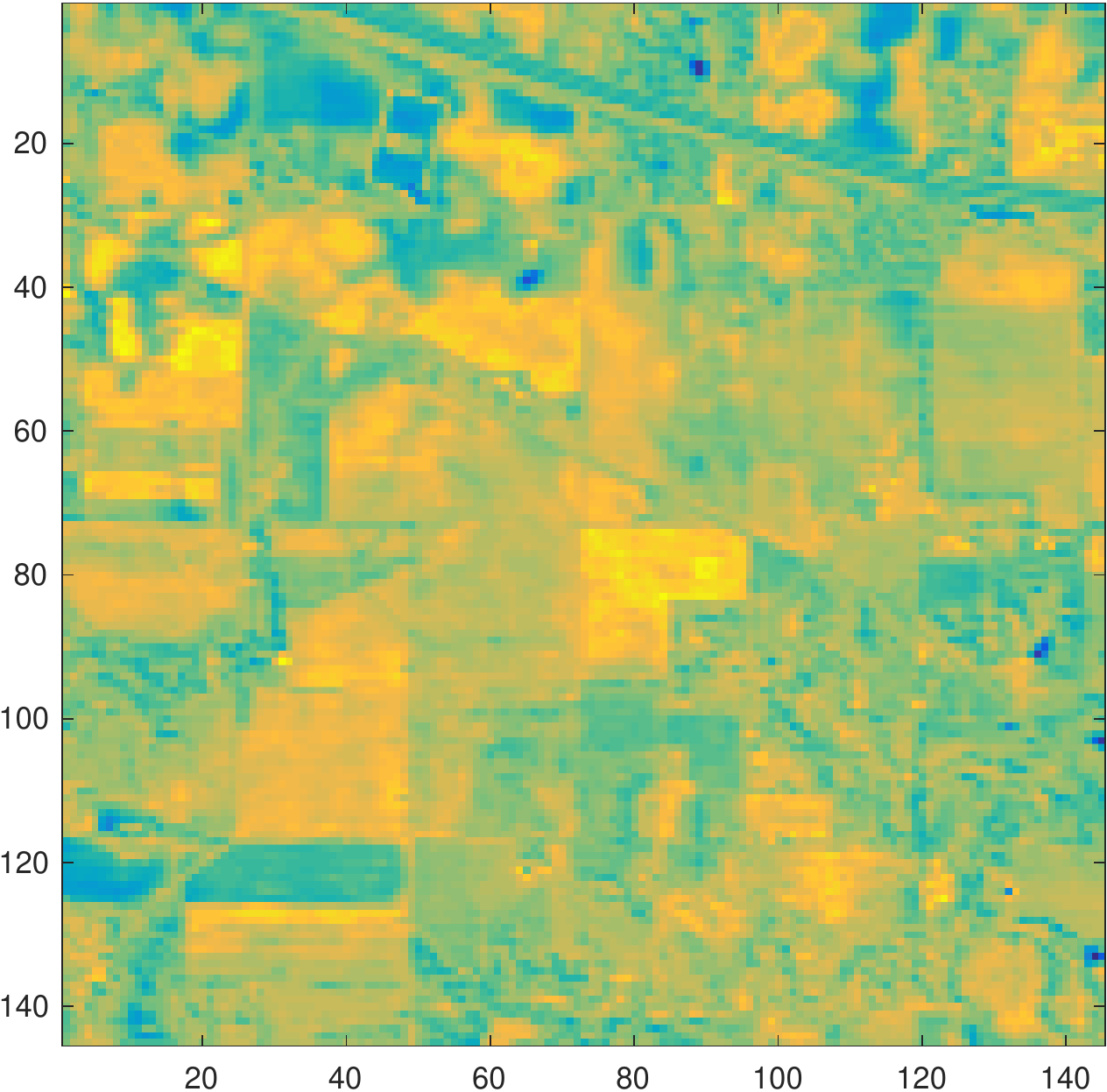,width=0.27\linewidth,clip=}&
   \epsfig{file=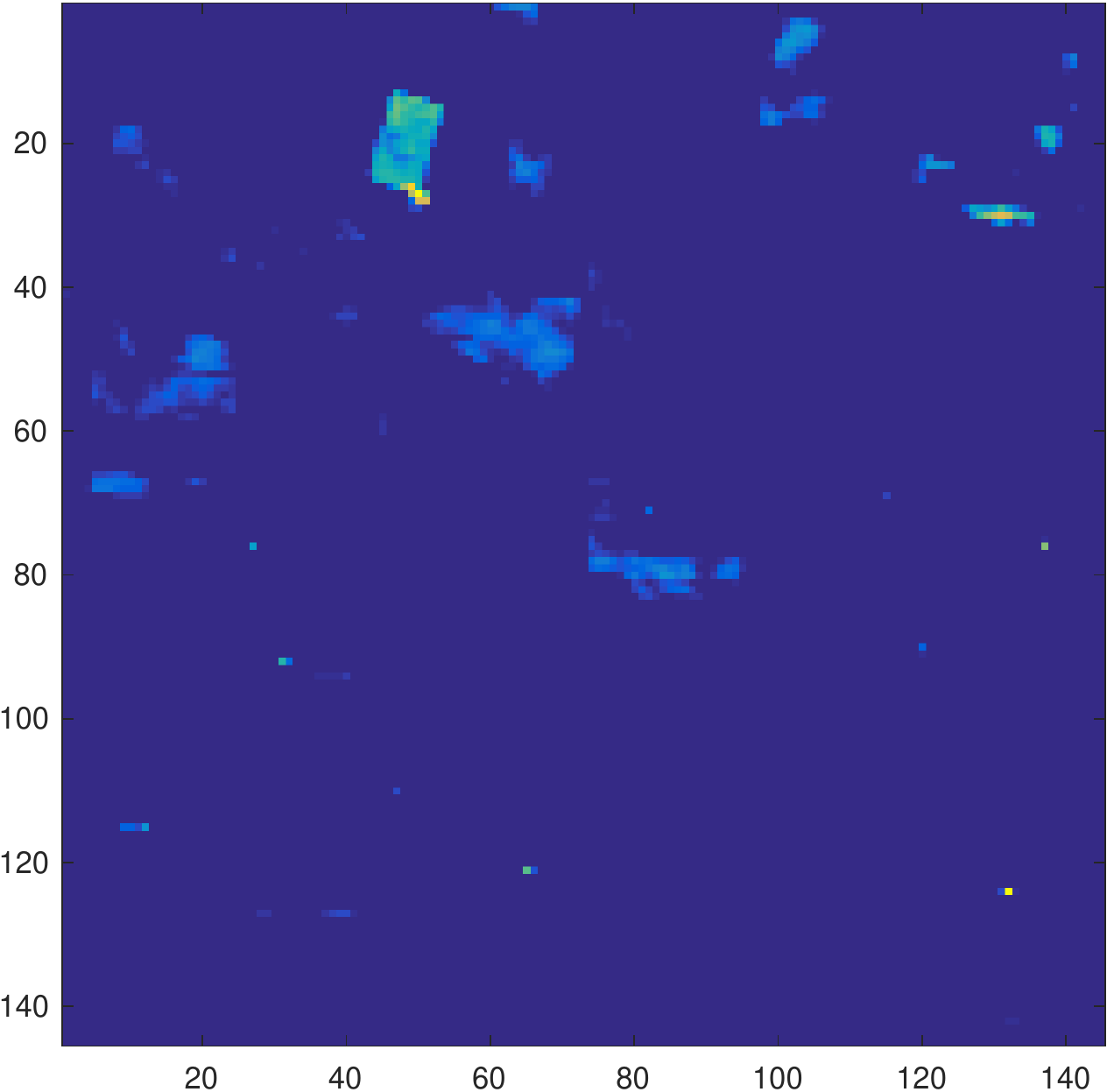,width=0.27\linewidth,clip=} \vspace{-5pt}\\ 
       (a) &&(c) & (d) \\
   \epsfig{file=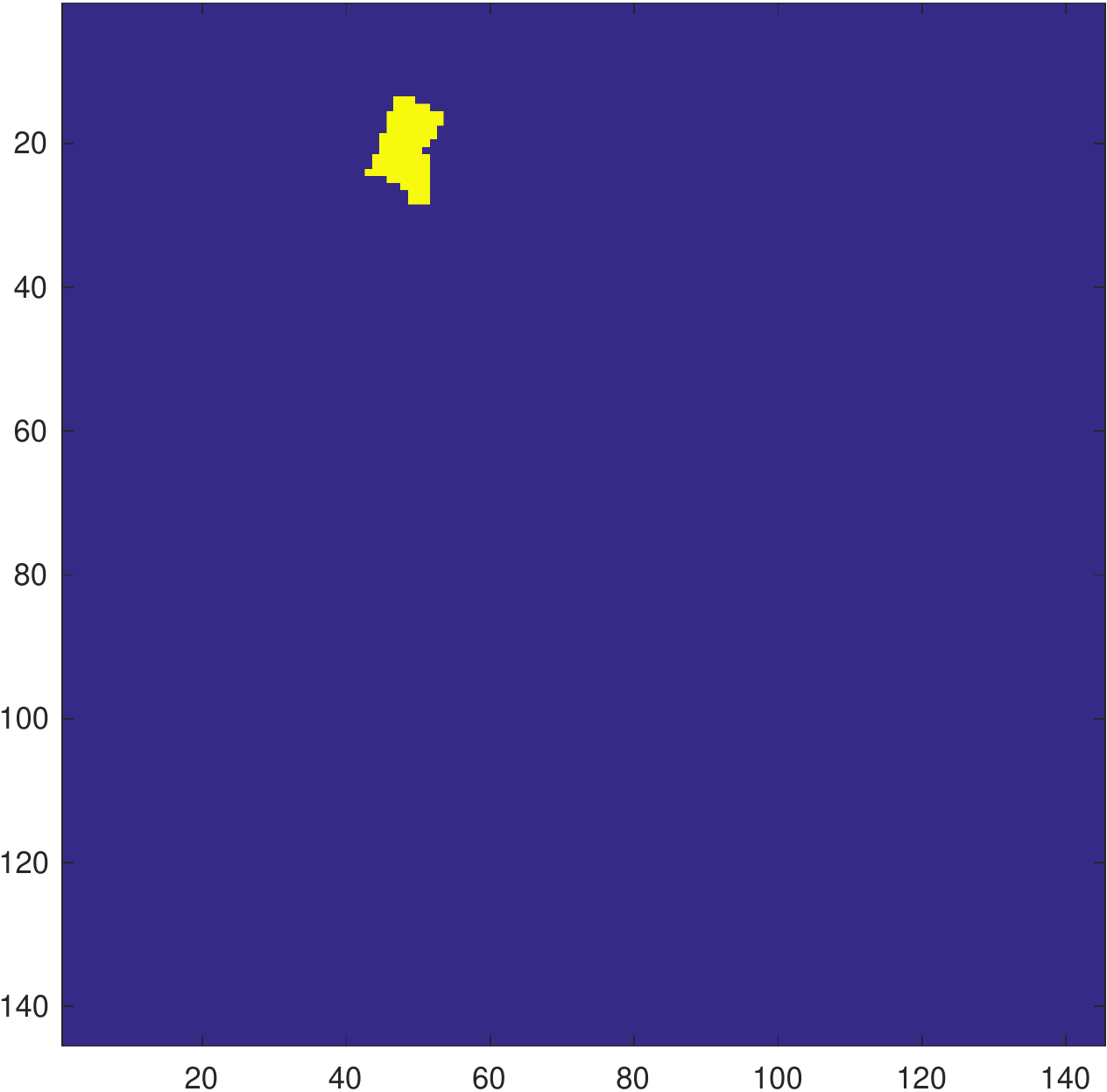,width=0.27\linewidth,clip=}&
   {\small\rotatebox{90}{ ~~~~{$85\%$ of $\lambda_{\text{max}}$}}} &
   \epsfig{file=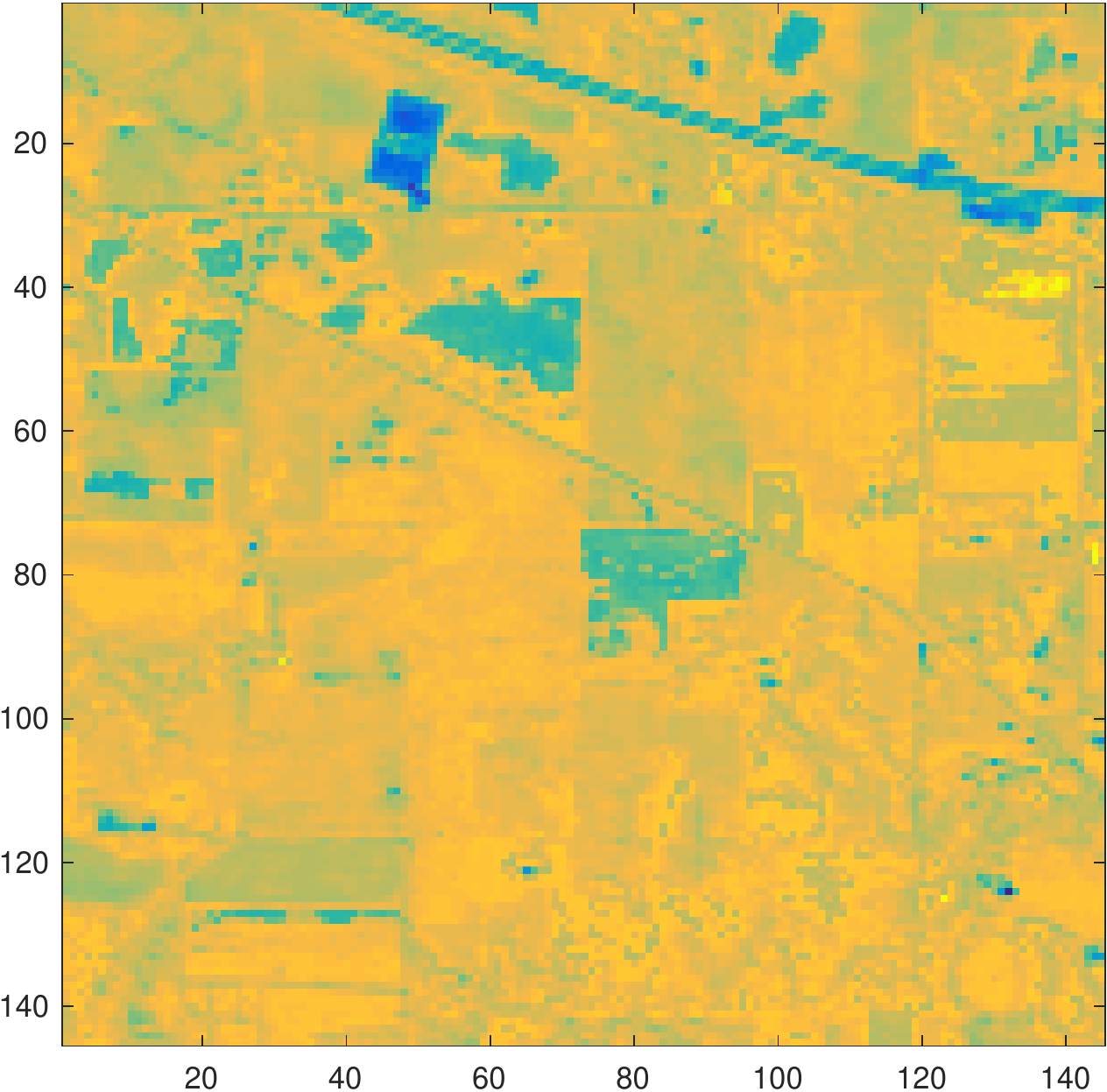,width=0.27\linewidth,clip=} &
   \epsfig{file=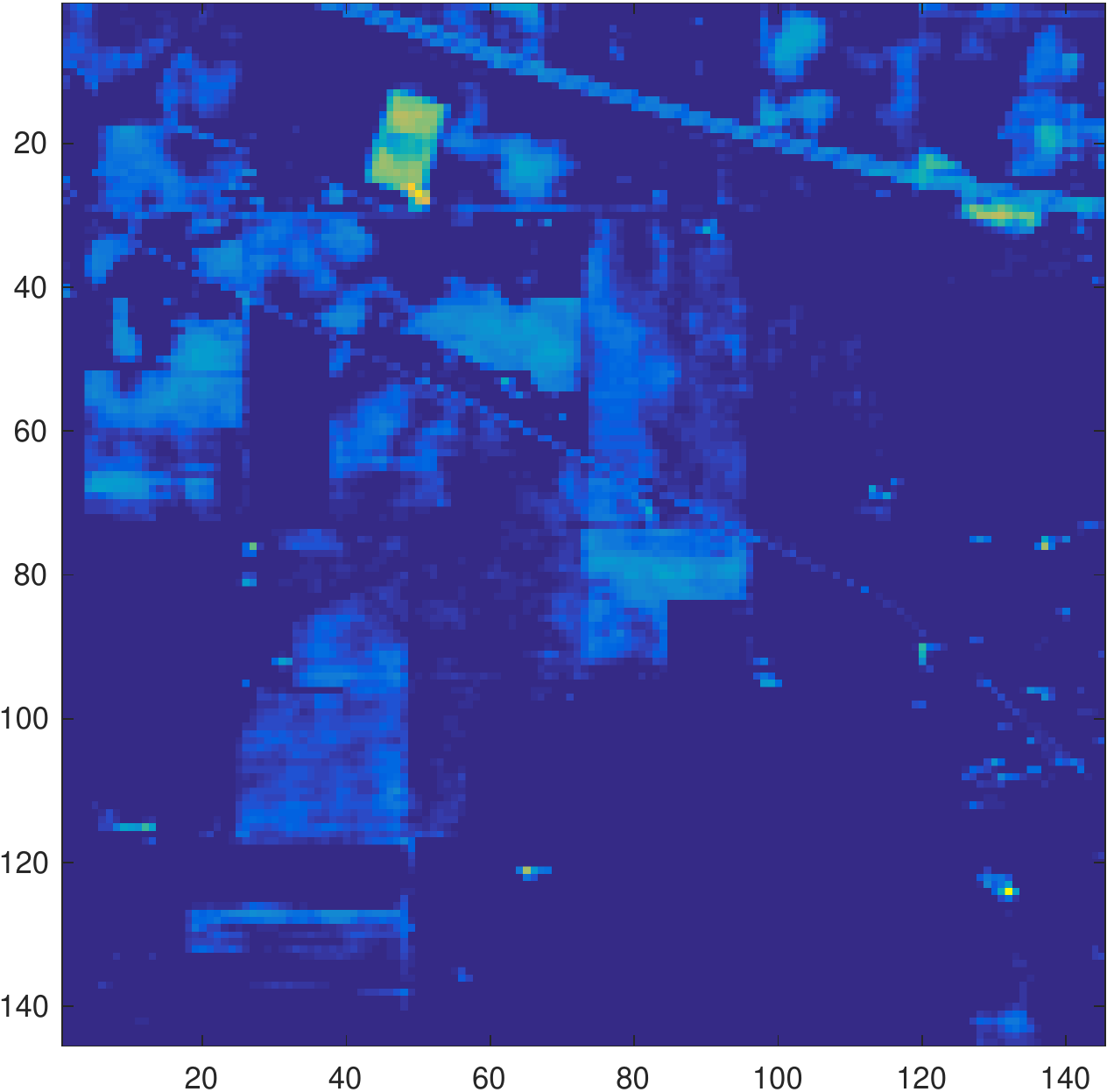,width=0.27\linewidth,clip=} \vspace{-5pt}\\ 
    (b)  && (e) & (f)
\end{tabular}
\caption{Recovery of the low-rank component {\small$\mb{X}$} and the dictionary sparse component {\small$\mb{RA}$} for different values of $\lambda$ for the proposed technique at {\small $f =50$} slice of the data (a), (b) corresponding ground truth for class-$16$. Panel (c) and (d),and (e) and (f) show the recovery of the the low-rank part and the dictionary part, for the $\lambda$ at the best operating point and for smaller $\lambda$, respectively, dictionary formed from data voxels(Table~\ref{res_tab}(c)).} 
\label{figure:res_our}
\end{figure}
 %
%

There are other interesting recovery results which warrant our attention. For example, Fig.~\ref{figure:res_our} shows the low-rank component and the dictionary sparse component recovered by the proposed method for two different values of $\lambda$, for the case where we form the dictionary by randomly sampling the voxels (Table~\ref{res_tab}(c)).  Interestingly, we recover the rail tracks/roads running diagonally on the top-right corner, along with some low-density housing; See Fig~\ref{figure:res_our} (f). This is because the \textit{signatures} we seek (stone-steel towers) are similar to those of the materials used in these structures. This further corroborates the applicability of the proposed approach in detecting the presence of a particular spectral \textit{signature} in a HS image. However, this also highlights potential drawback of this technique. As \texttt{XpRA} is based on identifying materials with similar composition, it may not be effective in distinguishing between very closely related classes, say two agricultural crops.
\section{Conclusions}
\label{sec:conclusion}
 We present a generalized robust PCA-based technique to localize a target in a HS image, based on the \textit{a priori} known spectral \textit{signature} of the material we wish to localize. We model the data as being composed of a low-rank component and a dictionary sparse component. Here, the dictionary contains the \textit{a priori} known spectral \textit{signatures} of the target. In this work, we adapt the theoretical results of \cite{Rambhatla2016} and present the conditions under which such a decomposition recovers the two components for the HS demixing task. Further, we evaluate and compare the performance of the proposed method via experimental evaluations on real HS image data for a classification task.
 \section{Acknowledgement}
 The authors graciously acknowledge support from the DARPA YFA, Grant N66001-14-1-4047.
\bibliographystyle{IEEEbib}
\bibliography{referLR}

\end{document}